\documentclass[conference]{IEEEtran}
\IEEEoverridecommandlockouts

\usepackage{cite}
\usepackage{amsmath,amssymb,amsfonts}
\usepackage{algorithmic}
\usepackage{graphicx}
\usepackage{textcomp}
\usepackage{xcolor}

\usepackage{subfigure}
\usepackage[ruled]{algorithm2e}

\begin{document}

\title{Identifying Training Stop Point with Noisy Labeled Data\\

\thanks{\copyright 2020 IEEE.  Personal use of this material is permitted. Permission from IEEE must be obtained for all other uses, in any current or future media, including reprinting/republishing this material for advertising or promotional purposes, creating new collective works, for resale or redistribution to servers or lists, or reuse of any copyrighted component of this work in other works.}
}

\author{\IEEEauthorblockN{Sree Ram Kamabattula and Venkat Devarajan}
\IEEEauthorblockA{\textit{Department of Electrical Engineering} \\
The Univeristy of Texas at Arlington, USA\\
sreeram.kamabattula@mavs.uta.edu\\
venkat@uta.edu
}
\and
\IEEEauthorblockN{Babak Namazi and Ganesh Sankaranarayanan}
\IEEEauthorblockA{\textit{Baylor Scott and White Health, Dallas, USA} \\
babak.namazi@bswhealth.org\\
ganesh.sankaranarayanan@bswhealth.org
}
}

\maketitle

\begin{abstract}

Training deep neural networks (DNNs) with noisy labels is a challenging problem due to over-parameterization. DNNs tend to essentially fit on clean samples at a higher rate in the initial stages, and later fit on the noisy samples at a relatively lower rate. Thus, with a noisy dataset, the test accuracy increases initially and drops in the later stages. To find an early stopping point at the maximum obtainable test accuracy (MOTA), recent studies assume either that i) a clean validation set is available or ii) the noise ratio is known, or, both. However, often a clean validation set is unavailable, and the noise estimation can be inaccurate. To overcome these issues, we provide a novel training solution, free of these conditions. We analyze the rate of change of the training accuracy for different noise ratios under different conditions to identify a training stop region. We further develop a heuristic algorithm based on a small-learning assumption to find a training stop point (TSP) at or close to MOTA. To the best of our knowledge, our method is the first to rely solely on the \textit{training behavior}, while utilizing the entire training set, to automatically find a TSP. We validated the robustness of our algorithm (AutoTSP) through several experiments on CIFAR-10, CIFAR-100, and a real-world noisy dataset for different noise ratios, noise types and architectures. 


\end{abstract}


\section{Introduction}
\label{sec:introduction}
DNNs have achieved remarkable performance in several computer vision domains due to the availability of large datasets and advanced deep learning techniques. However, in areas such as medical imaging, vast amount of data is still unlabeled. Labeling by experts is time-consuming and expensive, and is often performed by crowd sourcing \cite{Yan2014LearningExpertise}, online queries \cite{Blum2003Noise-tolerantModel}, etc. Such datasets introduce a large number of noisy labels, caused by errors during the labeling process. This work focuses on label noise. Therefore, noise and label noise are used interchangeably throughout the paper.

Training DNNs robustly with noisy datasets has become challenging and prominent work \cite {Algan2019ImageSurvey, Xiao2015LearningClassification}. While DNNs are robust to a certain amount of noise \cite{Rolnick2017DeepNoise}, they have the high capacity to fit on the noisy samples \cite{Zhang2017UnderstandingGeneralization, Arpit2017ANetworks}. However, the performance of the DNNs declines as they begin to learn on a significant number of noisy samples.

One notable approach to deal with this problem is to select and train on only clean samples from the noisy training data. In \cite{Chang2017ActiveSamples}, authors check for inconsistency in predictions. O2UNet \cite{Huang2019O2U-Net:Networks} uses cyclic learning rate and gathers loss statistics to select clean samples. SELF \cite{Nguyen2019Self:Self-ensembling} takes ensemble of predictions at different epochs. 

Few methods utilize two networks to select clean samples. MentorNet \cite{Jiang2018MentorNet:Labels} pre-trains an extra network to supervise a StudentNet. Decoupling \cite{Malach2017DecouplingUpdate} trains two networks simultaneously, and the network parameters are updated only on the examples with different predictions. Co-teaching \cite{Han2018Co-teaching:Labels} updates one network's parameters based on the peer network's small-loss samples. Co-teaching is further improved in co-teaching+ \cite{Yu2019HowCorruption}. 

\emph{Small-loss observation} suggests that clean samples have smaller loss compared to noisy samples in the early stages of training. However, in the later stages, it is harder to distinguish clean and noisy samples based on the loss distribution. Therefore, it is desirable to stop the training in the initial stages. In \cite{Li2019GradientNetworks}, authors theoretically show that the DNNs can be robust to noise despite over-parameterization with early-stopping, when trained with a first order derivative-based optimization method.

However, identifying the point, where the network starts to fit on noisy samples, is a challenging problem. Most previous methods assume that either i) the noise ratio is known or ii) the noise type is known. For instance, when the noise ratio is unknown, in \cite{Chen2019UnderstandingLabels}, authors propose a cross-validation approach to estimate the noise based on the validation accuracy, which becomes inaccurate for higher number of classes and harder datasets. In \cite{Song2019Prestopping:Noise}, the authors assume that a clean validation set is available to find the early stopping point. However, often, a clean validation set is not available. In any case, \cite{Sun2019LimitedLabels} suggests that the overall performance of the network is sensitive to the validation set quality. Therefore, a clean or noisy validation set is not reliable to identify the best stop point.

On the other hand, we attempt to stop the training before the network learns on a significant amount of noisy samples without any of the above assumptions. We propose a novel heuristic approach called AutoTSP to identify a TSP by monitoring the \textit{training behavior} of the DNN. Briefly, our main contributions are:

\begin{itemize}

    \item We define different memorization stages using clean and noisy label recall to understand the learning behavior on noisy samples.

    \item We analyze the positive and negative rates of change of training accuracy to define a training stop region.
    
    \item We further develop a heuristic algorithm to identify the TSP that does not require a clean validation set, and is independent of noise estimation and noise type.
    
    \item We show that AutoTSP can be incorporated in other clean sample selection methods \cite{Chen2019UnderstandingLabels}, \cite{Han2018Co-teaching:Labels} to further improve the performance.

\end{itemize}

We validate AutoTSP on two clean benchmark datasets: CIFAR-10 and CIFAR-100. Since CIFAR-10 and CIFAR-100 are clean, we synthetically add noise by changing the true label $\hat{y}$ to a different label $y$ with a noise transition matrix $N_{ij}$. We employ two types of noise to evaluate our robustness: 1) Symmetric (Sym): The true label is changed to any other class symmetrically with rate $\tau$. ($N_{ij}=\frac{\tau}{c-1}$ for $i\neq j$). 2) Asymmetric (Asym): Noise is distributed with just one class, we change the true label to the label of the following class with rate $\tau$. Asymmetric datasets are more realistic, as the labeler is likely to confuse one class with another class. We also validate our algorithm on a real world noisy dataset ANIMAL-10N. We perform several experiments with different architectures: 9-layer CNN (CNN9), ResNet32 and ResNet110 \cite{He2016DeepRecognition} to show that AutoTSP is architecture independent. 

\section{Related work}

Several methods have been introduced to deal with the challenges of training with noisy labels. A detailed review of different methods is provide in \cite{Algan2019ImageSurvey}.

Few methods attempt to correct the noisy labels. Bootstrapping \cite{Reed2015TrainingBootstrapping} uses predictions of the network to correct the labels, but their method is prone to overfitting. Joint optimization \cite{Tanaka2018JointLabels} corrects the labels by updating parameters and labels alternately, but requires prior knowledge of noise distribution. SELFIE \cite{Song2019SELFIE:Learning} is a hybrid method, which selects the clean samples based on the small-loss observation and re-labels the noisy samples using the loss-correction method. D2L \cite{Ma2018Dimensionality-DrivenLabels} utilizes a local intrinsic dimensionality. Dividemix \cite{Li2020DIVIDEMIX:LEARNING} separates the training data into labeled and unlabeled based on the loss distribution and trained in a semi-supervised manner.

Another approach is to correct the loss function by estimating the noise transition matrix. \cite{Goldberger2017TRAININGLAYER} added an extra softmax layer, \cite{Patrini2017MakingApproach} introduce forward and backward loss. However, the noise estimation becomes inaccurate, especially when the number of classes or the number of noisy samples is large.

Other related methods assign higher weights to clean samples for effective training. \cite{Ren2018LearningLearning} assigned weights to training samples by minimizing the loss on a clean validation set. CleanNet \cite{Lee2018CleanNet:Noise} produces weights for each sample during network training but requires human supervision.

A few related methods assign higher weights to clean samples for effective training \cite{Ren2018LearningLearning}, \cite{Lee2018CleanNet:Noise}. Few other works develop noise-robust loss functions \cite{Wang2019IMAEMatters}, \cite{Zhang2018GeneralizedLabels}, \cite{Wang2019SymmetricLabels}

\section{Understanding training stop point}
\label{sec: Ideal time to stop training}

Ideally, to find an early stopping point, a clean validation set is required, which is not often available. Thus in this section, as an alternative to the ideal requirement, we analyze the training behavior of the test accuracy with respect to the training accuracy, when we possess the ground truth, i.e. information about clean and noisy labeled samples in the training dataset. We utilize this analysis on the training behavior to define a Training stop point (TSP) without the ground truth.

Let $E$ be the total number of epochs and $S$ denote the vector of number of correctly predicted samples at each epoch. Let $S^{(e)}$ represent the number of correctly predicted samples at epoch $e$ and $S_{clean}^{(e)}$ and $S_{noisy}^{(e)}$ be the number of clean and noisy samples in $S^{(e)}$. Let the training accuracy at each epoch be denoted by $y_{f}^{(e)} = \frac{S^{(e)}}{D}$, where D represents the number of training samples in the dataset. For each epoch, we define the label recall of the clean and noisy samples  as $LR_{clean}^{(e)} = \frac{S_{clean}^{(e)}}{D_{clean}}$ and $LR_{noisy}^{(e)} = \frac{S_{noisy}^{(e)}}{D_{noisy}}$ respectively, where $D_{clean}$ and $D_{noisy}$ represent the number of clean and noisy samples present in $D$. \textbf{$LR_{clean}$} and \textbf{$LR_{noisy}$} represent the vectors of label recall values at each epoch.

\begin{figure}[t]
\centering
\begin{subfigure}
  \centering
  \includegraphics[width= 0.255\textwidth]{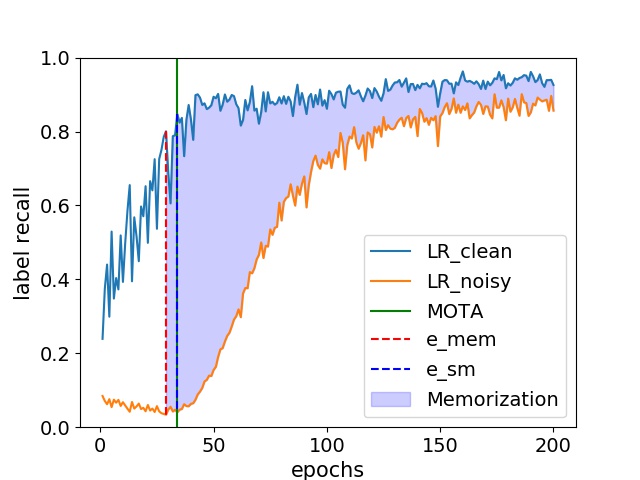}\includegraphics[width= 0.255\textwidth]{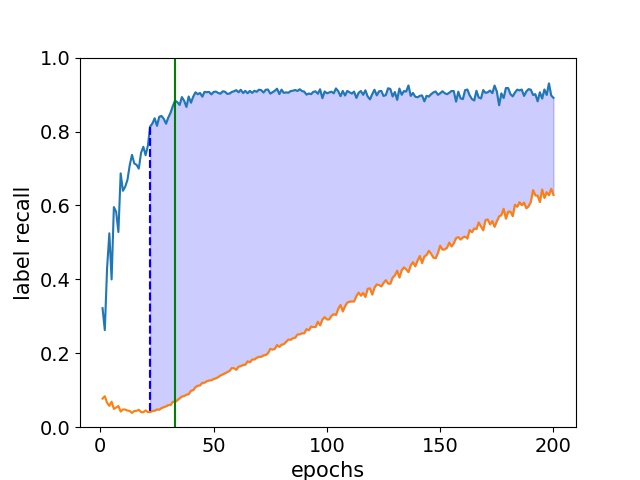}\\
    \vspace{-0.3\baselineskip}
   \includegraphics[width= 0.255\textwidth]{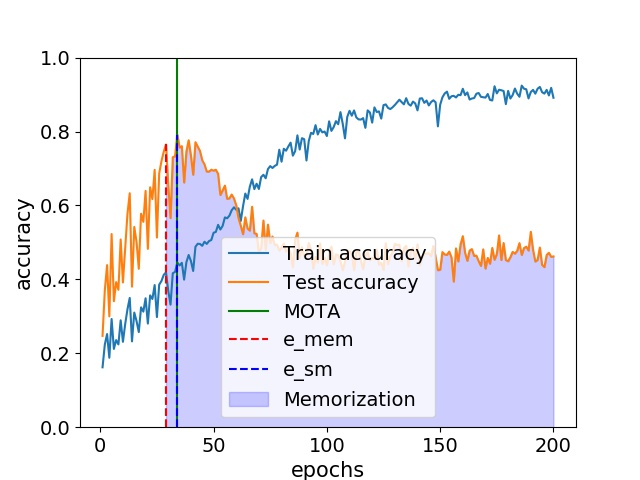}\includegraphics[width= 0.255\textwidth]{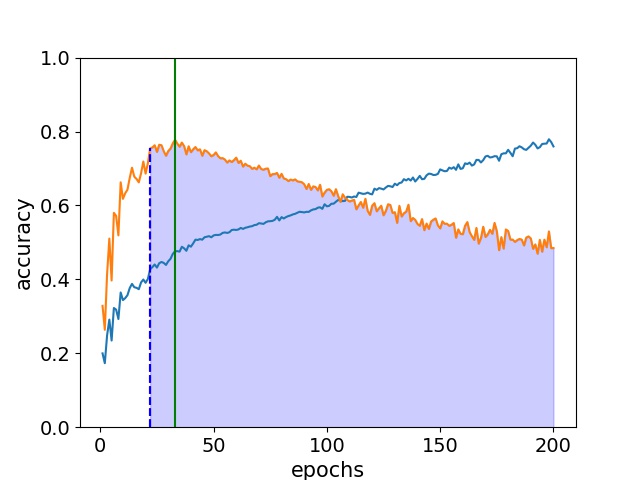}
  \caption{The memorization observation of CIFAR-10 0.5 Sym. Label recall on top figures, train and test accuracy on bottom figures with CNN9 (left) and  ResNet110 (right)}
  \label{fig: LR clean and noise}
\end{subfigure}
  \vspace{-1\baselineskip}
\end{figure}

We observe $LR_{clean}$ and $LR_{noisy}$ to understand the learning behavior of clean and noisy samples, i.e. to assess the region maximum obtainable test accuracy (MOTA) is achieved. In particular, 
we monitor the ratio of $\frac{LR_{clean}}{LR_{noisy}}$, and divide the training period into three memorization regions: pre-memorization ($PM$), mild-memorization ($MM$) and severe-memorization ($SM$). $LR_{clean}$ and $LR_{noisy}$ can be observed in the top plots of Fig.\ref{fig: LR clean and noise}, where the training accuracy and test accuracy are in the bottom plots.

We calculate the ratio of $\frac{LR_{clean}}{LR_{noisy}}$ and define two points: $e_{mem}$, the red vertical dotted line or the beginning of the shaded region and $e_{sm}$, the blue vertical dotted line as shown in Fig.\ref{fig: LR clean and noise}. 
 
The $e_{mem}$ separates the PM and MM regions by calculating the argmax of the ratio of label recall, i.e. the highest gap between the blue and the orange line as shown in the top plots of Fig.\ref{fig: LR clean and noise}. The $e_{sm}$ separates MM and SM regions by monitoring the point at which the ratio drops progressively. In some cases, with a high capacity network, the $e_{mem}$ and $e_{sm}$ might coincide, because the network can fit noisy examples at a high rate as shown in the right plot of Fig.\ref{fig: LR clean and noise}.

In the PM stage, before $e_{mem}$, the unshaded region in Fig.\ref{fig: LR clean and noise}, the learning is essentially on clean samples and thus the increase in the training accuracy is $proportional$ to the test accuracy. In particular, the $LR_{clean}$ increases at a higher rate, whereas the $LR_{noisy}$ remains minimal. In the MM stage (between $e_{mem}$ and $e_{sm}$), the rate at which $LR_{clean}$ increases drops and learning on noisy labels $LR_{noisy}$ increases gradually. Thus, the proportionality between training and test accuracy begins to attenuate. Later, in the SM region, as $LR_{noisy}$ keeps increasing, the $proportionality$ completely fails, as the learning is mostly on noisy labels. In other words, the test accuracy keeps dropping as the training accuracy increases. This can be seen in the bottom plots of Fig.\ref{fig: LR clean and noise}. The network learns on higher number of noisy samples in the SM stage, and it might not learn significant number of clean samples in the PM stage. Therefore, the desired training stopping point should be between $e_{mem}$ and $e_{sm}$.

However, $e_{mem}$ and $e_{sm}$ are obtained with the ground truth. Therefore, we utilize the learning behavior, specifically, the \textit{rate of change} of training accuracy at each memorization stage to find the desired TSP without the ground truth.

\section{Ideal time to stop training}
\label{Ideal time to stop training}

We monitor the rate of change of training accuracy $y_{f}^{(e)}$ at each epoch $e$, which can be either positive or negative. We represent the \emph{magnitude} of rate of change at epoch e by $pr^{(e)}$ if positive, or $nr^{(e)}$ if negative. We separate all the $pr^{(e)}$ and $nr^{(e)}$ into vectors \emph{PROCE} and \emph{NROCE} respectively. Now, we observe the variance of magnitudes of the \emph{PROCE} and \emph{NROCE} in different memorization stages.

In the PM stage, both $pr^{(e)}$ and $nr^{(e)}$ have high magnitude consistently. Thus, we can observe higher spikes in the unshaded region in Fig.\ref{fig: LR clean and noise}. Therefore, we assume there would be very few epochs with small-magnitude initially. We refer to these epochs with small-magnitude as small learning epochs (SLEs). As the network begins to fit on the noisy samples, in the MM region, the magnitudes of both $pr^{(e)}$ and $nr^{(e)}$ decrease, so there would be higher number of SLEs. We would also observe fewer $nr^{(e)}$ epochs, resulting in a smoothness in the training accuracy curve near the SM region. As the network severely memorizes the noisy labels in the SM region, there would still be higher number of SLEs. But, towards the end, as the network fits on all the training examples, the rate tends to increase slightly, causing the oscillation towards the end in Fig.\ref{fig: LR clean and noise}. The magnitude in the SM region varies with different settings, but, the drop in the magnitude near the SM region (smoothness) is consistent across different architectures, datasets, noise types and noise ratios as shown in later figures. From our analysis in the previous section, the desired training stop point would be between $e_{mem}$ and $e_{sm}$. Therefore, we identify this smoothness i.e. find the initial reduction in the \emph{PROCE} and \emph{NROCE} to choose a training stop region.

We divide the \emph{PROCE} and \emph{NROCE} into intervals of fixed length $\beta_1$ and calculate the cumulative sum of magnitudes in each interval. Then, we standardize (zero mean and unit variance) these sums and identify the interval, where the standardized value becomes less than zero. Two different reduction points will be obtained for PROCE and NROCE (labeled as posline and negline in figures). We define this region between the two points as the training stop region.

\begin{figure}[t]
\centering
\begin{subfigure}
  \centering
  \includegraphics[width= 0.255\textwidth]{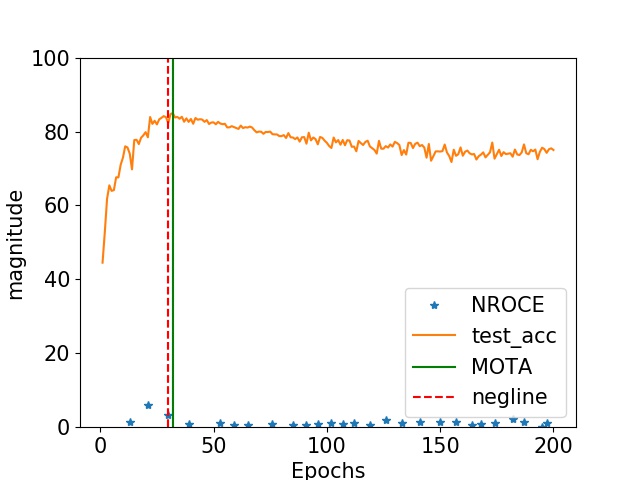}\includegraphics[width= 0.255\textwidth]{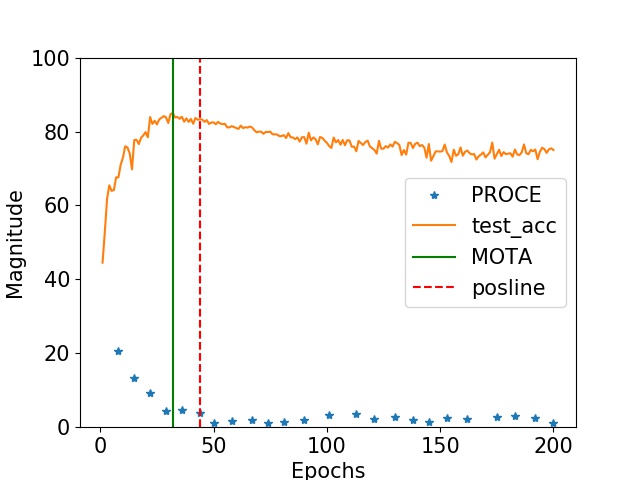}\\
  \vspace{-0.3\baselineskip}
   \includegraphics[width= 0.255\textwidth]{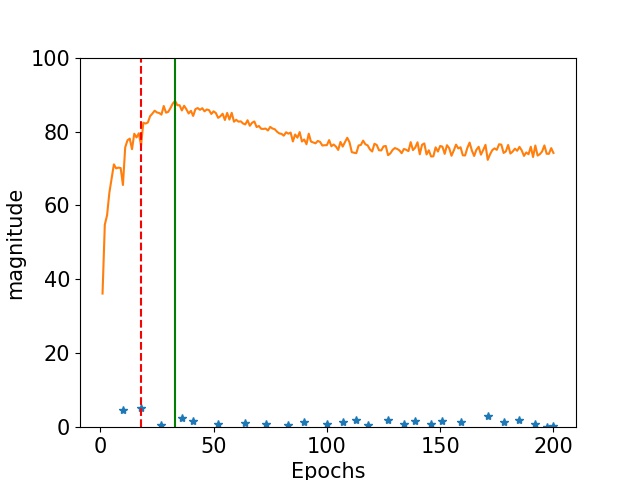}\includegraphics[width= 0.255\textwidth]{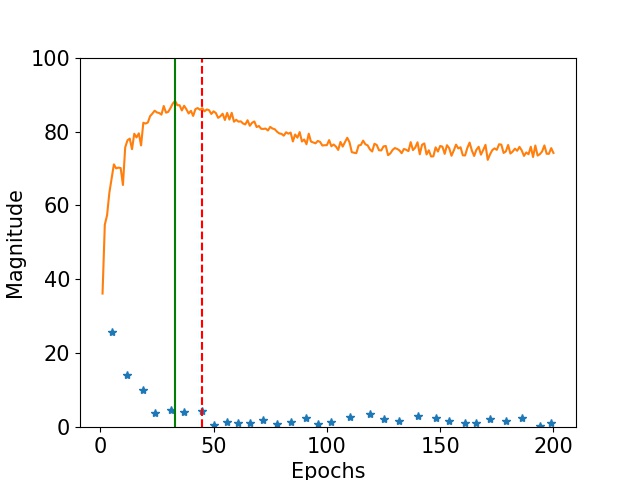}
  \caption{NROCE on left, PROCE on right with ResNet110 on CIFAR-10: 0.2 Sym on top, 0.2 Asym on bottom}
  \label{fig: NROC PROC CIFAR10 sym asym 0.2}
\end{subfigure}
  \vspace{-1\baselineskip}
\end{figure}

We plot the NROCE, PROCE along with the test accuracy in Fig.\ref{fig: NROC PROC CIFAR10 sym asym 0.2}. The blue asterisk points in the figure are located at the epoch numbers in NROCE and PROCE, spaced apart by length $\beta_1$. The height of the blue asterisk points represents the cumulative sum at each interval. The red vertical dotted line is plotted at the point, where the reduction happens.

The small-learning belief can be seen in Fig.\ref{fig: NROC PROC CIFAR10 sym asym 0.2}, i.e. the asterisk points are higher in the beginning and drop in the later stages. It can be observed that the red dotted line on both the figures are close to the MOTA. Therefore, this supports our training stop region consideration.

\begin{figure}[t]
\centering
\begin{subfigure}
  \centering
  \includegraphics[width= 0.255\textwidth]{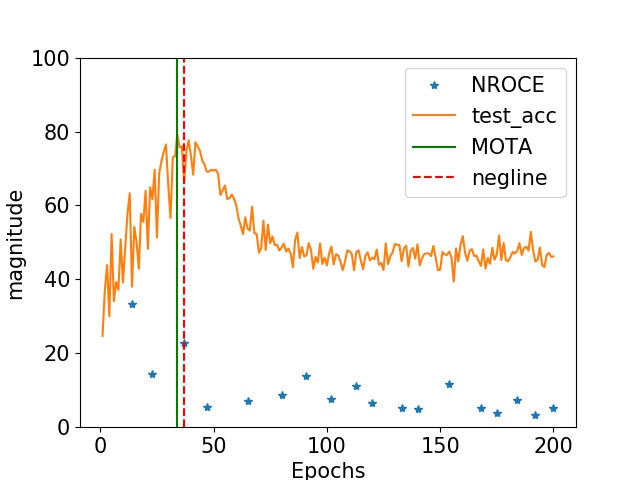}\includegraphics[width= 0.255\textwidth]{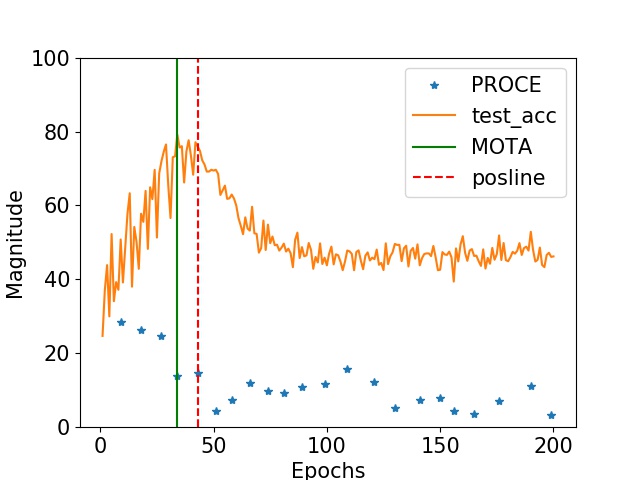}\\
      \vspace{-0.3\baselineskip}
   \includegraphics[width= 0.255\textwidth]{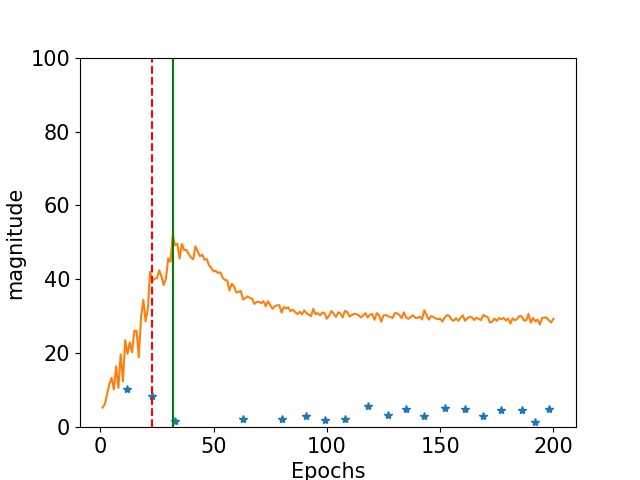}\includegraphics[width= 0.255\textwidth]{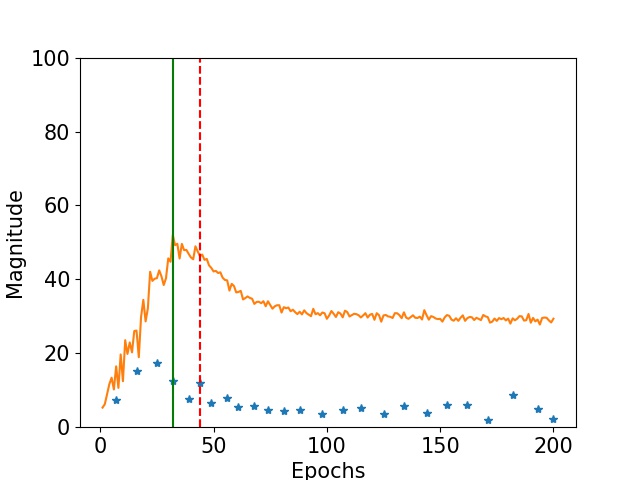}
  \caption{NROCE on left, PROCE on right with CNN9 0.5 Sym: CIFAR-10 on top, CIFAR-100 on bottom}
  \label{fig: Different architecture NROC PROC CIFAR10 and CIFAR100 sym 0.5}
\end{subfigure}
  \vspace{-1\baselineskip}

\end{figure}

Fig.\ref{fig: Different architecture NROC PROC CIFAR10 and CIFAR100 sym 0.5} shows plots for the higher noise ratio with a different architecture. The PROCE asterisk points are higher initially, but subsequently vary in the SM stage as discussed earlier. However, as the smoothness is consistent, the red dotted lines of both PROCE and NROCE are still close to MOTA.

\begin{figure}[t]
\centering
\begin{subfigure}
  \centering
  \includegraphics[width= 0.255\textwidth]{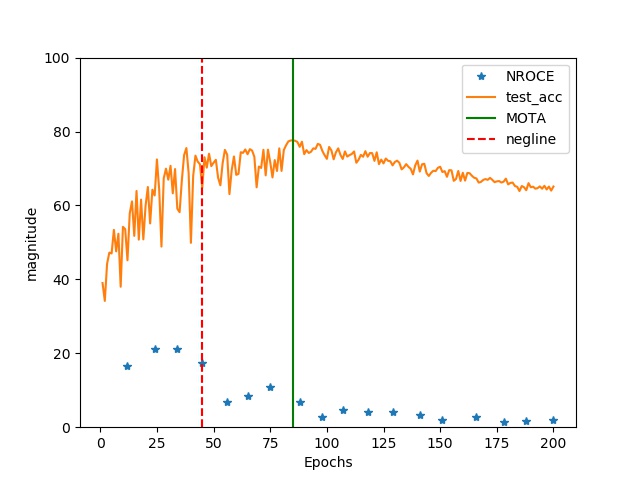}\includegraphics[width= 0.255\textwidth]{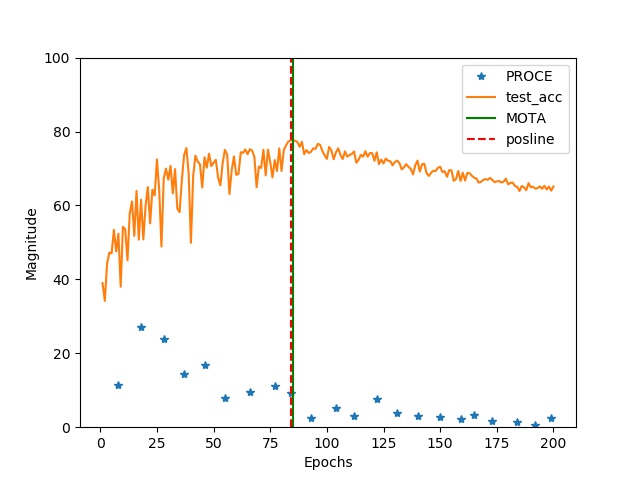}\\
      \vspace{-0.3\baselineskip}
   \includegraphics[width= 0.255\textwidth]{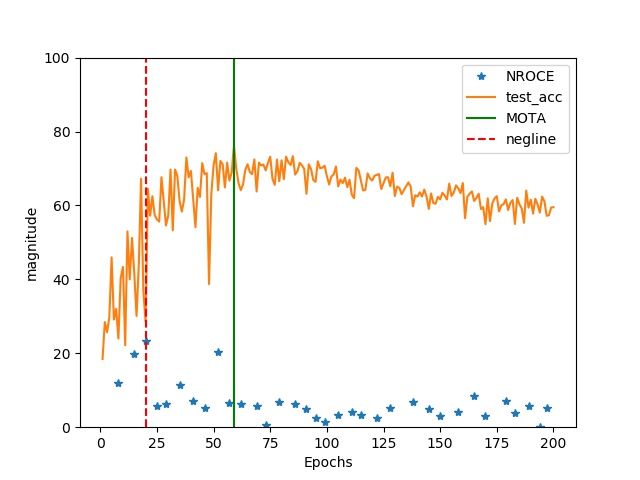}\includegraphics[width= 0.255\textwidth]{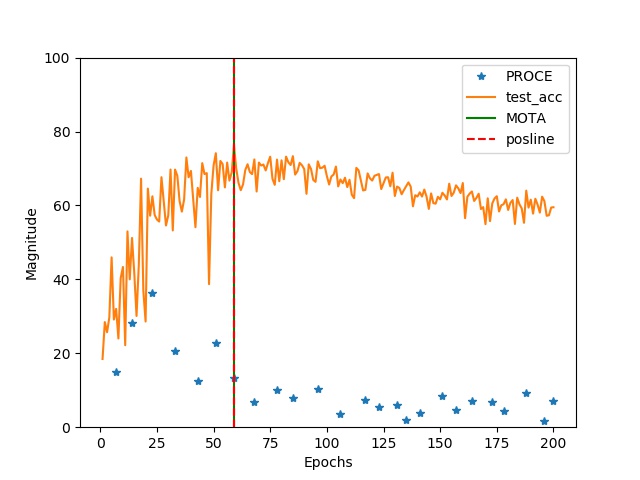}
  \caption{NROCE on left and PROCE on right when trained on CIFAR 10 with ResNet32 on 0.2 symmetric noise with LR2 on top, LR3 on bottom}
  \label{fig: Different learning rate architecture NROC PROC CIFAR10 and CIFAR100 sym 0.5}
\end{subfigure}
  \vspace{-1\baselineskip}

\end{figure}

Since, our belief depends on the rate of change of training accuracy, we check the sensitivity of our belief on different learning rate schedules. The top plot in Fig.\ref{fig: Different learning rate architecture NROC PROC CIFAR10 and CIFAR100 sym 0.5} are of 0.001 initial rate and multiplied by 0.5, 0.25, 0.1 at 80, 120, 160 epochs respectively (LR2). The bottom plot is of constant learning rate 0.001 (LR3). The reduction of sums in the later stages can still be observed in both cases. This further verifies the robustness of our training stop region.

Now, we develop a heuristic algorithm Alg. \ref{algortihrm: AutoTSP} to identify the training stop point (TSP) within the training stop region. We calculate the maximum training accuracy which stores $y_{f}^{(e)}$ only if $y_{f}^{(e)}$ is greater than $y_{f}^{(e-1)}$, else it retains $y_{f}^{(e-1)}$. We calculate the rate of change of this vector denoted by \emph{MT}, where $MT^{(e)}$ represents the magnitude at epoch $e$. As discussed earlier, there would be very few SLEs in $MT$ initially. In contrast, there would be higher number of SLEs in the later stages. Since, the test accuracy decreases in the later stages, we assume that the network learns on the noisy samples at SLEs, which results in a drop in test accuracy. We define an epoch as a small-learning epoch (SLE), if the magnitude value $MT^{(e)}$ is less than the threshold value $\theta_{1}$. We negate the values of these SLEs in $MT$ to indicate that the test accuracy is dropped at these epochs. To identify the TSP, we determine whether the epochs following the SLEs compensate the drop in the test accuracy at SLEs, and achieve higher test accuracy. Therefore, we define a vector $C$ to calculate the cumulative sum of the SLEs and the following epochs until the sum becomes positive, which we believe indicates the increase in test accuracy. We store the cumulative sum at epoch e in $C^{(e)}$. We set the $C^{(e)}$ to zero, if the sum at epoch $e$ is negative, to indicate that the test accuracy is not increasing. Now, we calculate the sums of consecutive values in $C$ within the training stop region and store them in vector $V$. We consider the highest cumulative sum point (argmax) in $V$ as the training stop point $e_{tsp}$.

\begin{algorithm}[t]
\SetAlgoLined 
\textbf{Input:} {$\beta_{1}$,$nr$,$pr$,$E$,$MT$, $\theta_{1}$, $\theta_{2}$ }\\
\For{$e \leftarrow 0$ \textbf{to} $E-1$} 
{$N_{cum}^{(e/\beta_1)} \leftarrow sum(nr^{(e)} \textbf{until} nr^{(e+\beta1-1)}$) \\
$P_{cum}^{(e/\beta_1)} \leftarrow sum(pr^{(e)} \textbf{until} pr^{(e+\beta_{1}-1)}$) \\
$e \leftarrow e+\beta_{1}$}
Then \textbf{standardize} $N_{cum}$, $P_{cum}$ $\rightarrow$  $N_{stand}$, $P_{stand}$\\
\textbf{Pass} through $N_{stand}$ and $P_{stand}$ /*monitoring the rate of change*/\\
\If{$N_{stand}^{(l)}<0$}{$e_{nroce}$  = l*$\beta_1$ \\/* Similarly find $e_{proce}$ for $P_{stand}$*/}
/*Training stop region: [min($e_{nroce}$, $e_{proce}$), max($e_{nroce}$, $e_{proce}$)]*/\\
\textbf{set} $init$ to 0\\
\For{$e \leftarrow 0$ \textbf{to} $E-1$}
{\If{($MT^{(e)}>\theta_{1}$)}{$C^{(e)}$ = $MT^{(e)}$ + $init$
\textbf{and} \textbf{set} $init$ to 0}
\Else{$C^{(e)}$ = 0 \textbf{and} \textbf{set} $init$ = \textbf{-}$MT^{(e)}$}
\If{$MT^{(e)}<\theta_{2}$}{\textbf{then} $C^{(e)}$ =0}}
V $\leftarrow $ Sums of consecutive non-zero elements of $C$ within the training stop region.\\
\textbf{/*Training stop point*/} $e_{tsp}$ = argmax($V$)

 \caption{AutoTSP}
 \label{algortihrm: AutoTSP}
 \textbf{Output:} {$e_{tsp}$}
\end{algorithm}

\section{Experiments}

 We use the Adam optimizer and a momentum of 0.9. The batch size is set to 128 and the network is trained for 200 epochs. The initial learning rate is set to 0.001 and multiplied by 0.5, 0.25, 0.1 at 20, 30 and 40 epochs respectively for all experiments, otherwise mentioned. We set the algorithm hyper-parameters $\beta_{1}$ to 5, 6 and 7 denoted by \{5,6,7\}, and $\theta_{1}$, $\theta_{2}$ to 0.5 based on experimental observations.

\subsection{Performance Comparison of AutoTSP:}
\label{Performance Comparison of early stopping point}

In this section, we compare the AutoTSP when all else are equal: use of entire training data, assumption that a clean validation set is not available. Thus, we compare our method only with 1) MOTA, point where the maximum test accuracy is obtained, as represented in the figures by the green vertical line. 2) Standard training, i.e., test accuracy at the end of the epochs without early stopping (labeled as Standard in the tables). 3) Additionally, we also compare with Noise Heuristic Accuracy (NHA) \cite{Song2019Prestopping:Noise}, which assumes that the noise ratio $\tau$ is known, an assumption, which we believe should be avoided. NHA suggests that during training on a $\tau$ percent noisy dataset, the best point to stop is when the training accuracy reaches $(1-\tau)$ percent.

We measure the performance of AutoTSP to the above-mentioned methods with test accuracy (acc), label precision (LP) and label recall (LR) of the training data ($LP = \frac{S_{clean}^{(e_{tsp})}}{S^{(e_{tsp})}}$, $LR = \frac{S_{clean}^{(e_{tsp})}}{D_{clean}}$).

Table \ref{TSP comparison ResNet110} reports the results on ResNet110 architecture. It can be seen that the AutoTSP is either exact or close to MOTA across different noise ratios, datasets and noise types. It can also be observed that AutoTSP outperforms NHA in most of the cases. The results also confirm that the standard training performs poorly when training with noisy labels.

\begin{table}[t]
\caption{Comparison of results with ResNet110 }
    \centering
    \resizebox{\columnwidth}{!}{%

    \begin{tabular}{|c|c|c|c|c|c|c|c|c|}
    \hline
     & \multicolumn{4}{|c|}{\textbf{CIFAR 10}} & \multicolumn{4}{|c|}{\textbf{CIFAR 100}}\\
     & \multicolumn{2}{|c|}{Symmetric} & \multicolumn{2}{|c|}{Asymmetric}& \multicolumn{2}{|c|}{Symmetric} & \multicolumn{2}{|c|}{Asymmetric}\\
     &0.2&0.5&0.2&0.4&0.2&0.5&0.2&0.4\\
    \hline
    \hline
        MOTA (acc) &85.01 &75.86&88.28&77.78&58.96&46.65&57.99&43.7 \\
       
        AutoTSP (acc) & \textbf{85.01}&\textbf{75.86}&\textbf{88.28}&\textbf{76.31}&\textbf{57.33}&\textbf{46.65}&56.03&\textbf{42.25}\\
        NHA (acc) & 83.19&70.7&84.22&73.57&56.35&40&\textbf{56.43}&40.25\\
        Standard (acc) & 75&48.78&74.20&56.67&52.09&27.33&53.98&39.66\\
        \hline
        MOTA (LP) & 97.6 & 95.2 &97.54& 87.94 & 99.14 & 96.45 &92.76& 71.81\\
        AutoTSP (LP) & 97.6 & 95.2 &97.54& 87.54 & 94.5 & 96.45 &84.25& 63.1\\ 
        NHA (LP) & 95.32 & 85.9 &95.18& 82.39 & 90.38 & 76.41 &87.24& 66.94\\ 
        Standard (LP) & 81.78 & 54.52 &80.92& 61.35 & 80.53 & 50.6 &79.84& 59.86\\ 
 \hline
 MOTA (LR)  & 94.23 & 80.23 &96.56& 87.62 & 76.07 & 61.48 &78.77& 64.96 \\
  AutoTSP (LR)  & 94.23 & 80.23 &96.56& 82.66 & 88.9 & 61.48 &93.83& 88.21 \\
   NHA (LR)  & 96.54 & 89.03 &94.41& 85.10 & 93.71 & 84.02 &89.46& 67.32 \\
    Standard (LR)  & 98.18 & 95.02 &97.5& 96.33 & 98.58 & 96.4 &99.12& 98.74 \\
    \hline
    \end{tabular}
    }
    \label{TSP comparison ResNet110}
\end{table}

\begin{figure}[t]
\centering
\begin{subfigure}
  \centering
   \includegraphics[width= 0.255\textwidth]{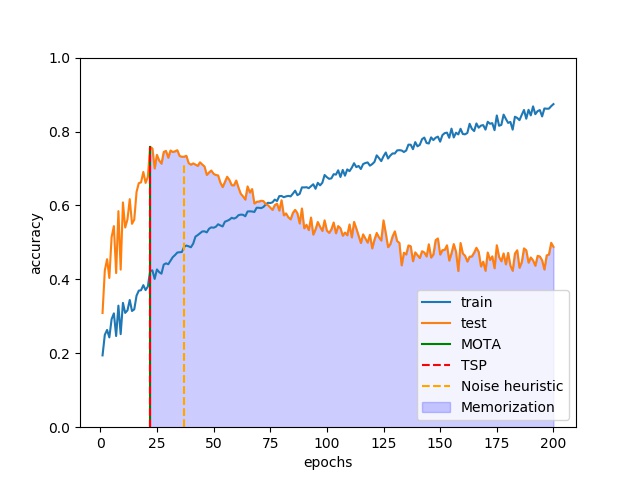}\includegraphics[width= 0.255\textwidth]{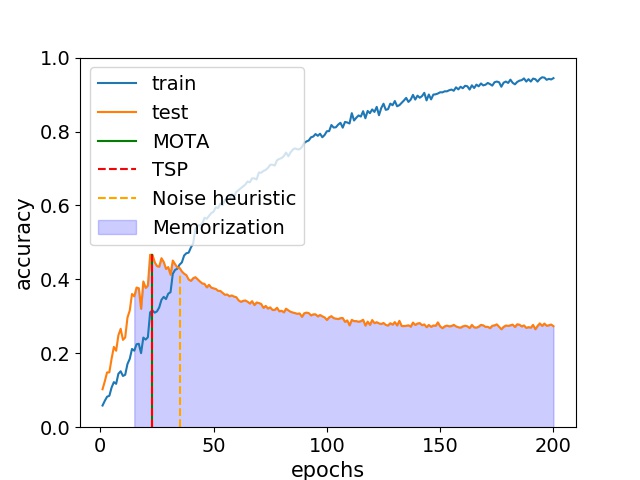}\\    
   \vspace{-0.3\baselineskip}
  \includegraphics[width= 0.255\textwidth]{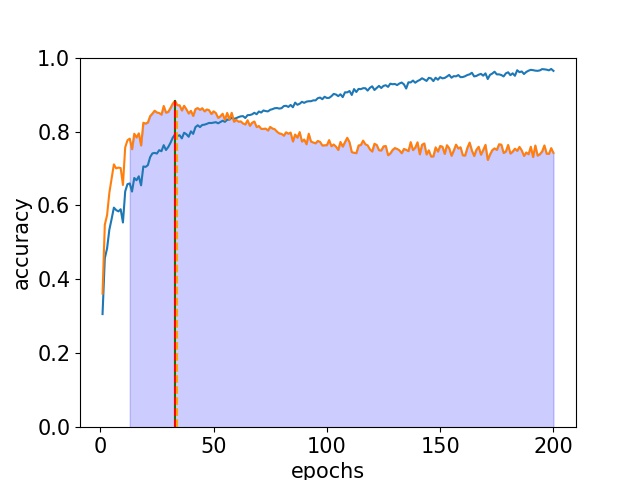}\includegraphics[width= 0.255\textwidth]{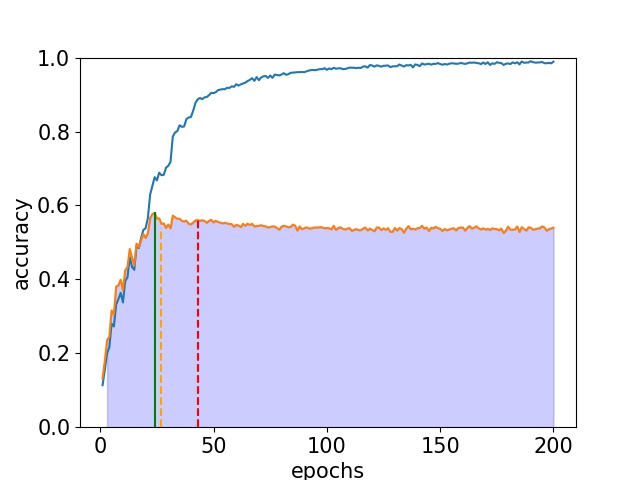}\\
  
  \caption{AutoTSP comparison with ResNet110: 0.5 Sym on top, 0.2 Asym on bottom, CIFAR-10 on left, CIFAR-100 on right}
  \label{fig: experiments refer1}
\end{subfigure}
  \vspace{-1\baselineskip}

\end{figure}

On CIFAR-10 dataset, with the easier 0.2 Sym noise, it can be seen that the AutoTSP is able to exactly identify the MOTA point. High LP indicates that the learning on noisy samples is minimal, and high LR represents that the network learned significant number of clean samples at the TSP. The significance of AutoTSP can be observed for the harder cases. The AutoTSP still finds the exact stop point at MOTA, when the noise ratio is increased to 0.5, as shown in the top left plot of Fig. \ref{fig: experiments refer1}. It performs similarly for both asymmetric cases as shown in the bottom left plot. On the other hand, NHA and standard achieve higher LR than the MOTA, because the network stops the training in the later stages, where the number of clean samples learned is higher. But, lower LP indicates that they learn on higher number of noisy samples due to the stop point in the SM region, which is not desired.

Results on the harder dataset CIFAR-100 are similar to CIFAR-10 for Sym as shown in the top righ plot of Fig.\ref{fig: experiments refer1}. For the Asym 0.2 and 0.4 cases, both AutoTSP and NHA are close to MOTA, as shown in the bottom right plot of Fig.\ref{fig: experiments refer1}. In this case, the LP and LR of AutoTSP are not as desired, because the network learns on noisy labels from the beginning of the training.

\begin{figure}[b]
\centering
\begin{subfigure}
  \centering
  \includegraphics[width= 0.23\textwidth]{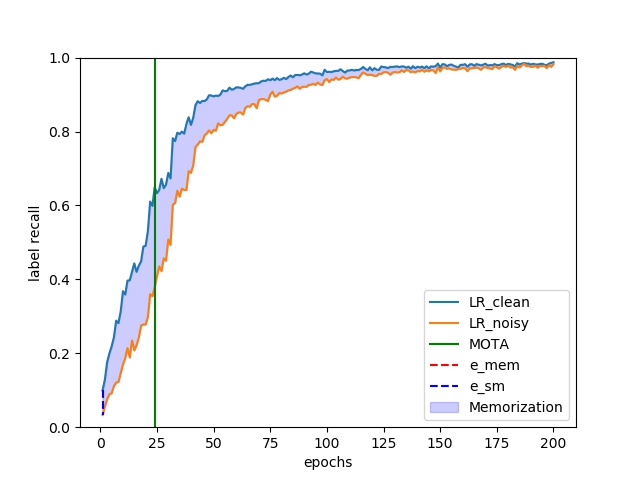}
    \includegraphics[width= 0.23\textwidth]{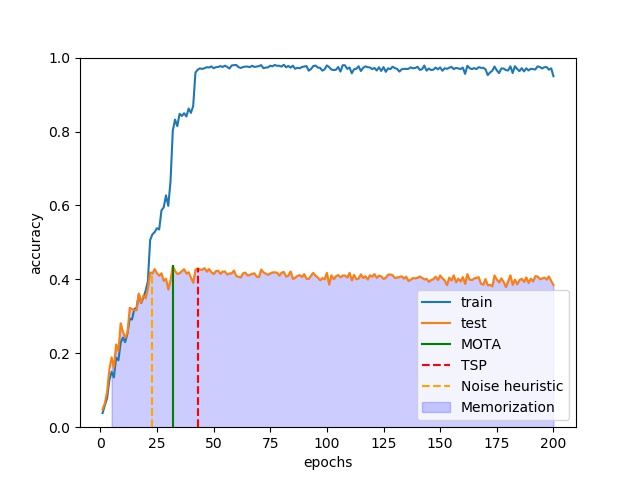}

  \caption{LR (left) and test accuracy (right) on CIFAR-100 with 0.4 Asym on ResNet110}
  \label{fig: asym label recall}
\end{subfigure}
\end{figure}

For harder asymmetric noise cases, the network learns on noisy examples significantly along with clean examples from the initial stages of training as shown in the left plot of Fig.\ref{fig: asym label recall}. It can be observed that the $LR_{noisy}$ (orange line) increases along with the $LR_{clean}$ (blue line). This suggests that the small-loss observation is not stronger in this case. Thus, the test accuracy remains the same even in the later stages. Surprisingly, in this case, NHA stops early than AutoTSP. However, the AutoTSP is still close to MOTA, as shown in the right plot of Fig. \ref{fig: asym label recall}. But it stops at a point later in the $SM$ region, resulting in high LR and low LP as shown in table \ref{TSP comparison ResNet110}.

We validated our algorithm with CNN9 as shown in Table \ref{TSP comparison CNN9}. It can be observed that the AutoTSP is still close to MOTA across different cases. As shown in Fig. \ref{fig: Different architecture NROC PROC CIFAR10 and CIFAR100 sym 0.5} bottom plot, the AutoTSP identifies a training stop region close to MOTA. However, AutoTSP finds a stop point in the beginning of the region, resulting in a low LR and higher LP. This is likely due to the hyper-parameter (argmax) selection of the stop point. 

\begin{table}[t]
\caption{Comparison of results with CNN9 }
    \centering
    \resizebox{\columnwidth}{!}{%

    \begin{tabular}{|c|c|c|c|c|c|c|c|c|}
    \hline
     & \multicolumn{4}{|c|}{\textbf{CIFAR 10}} & \multicolumn{4}{|c|}{\textbf{CIFAR 100}}\\
     & \multicolumn{2}{|c|}{Symmetric} & \multicolumn{2}{|c|}{Asymmetric}& \multicolumn{2}{|c|}{Symmetric} & \multicolumn{2}{|c|}{Asymmetric}\\
     &0.2&0.5&0.2&0.4&0.2&0.5&0.2&0.4\\
    \hline
    \hline
        MOTA (acc) &85.69 &79.19&87.58&81.82&60.84&52.03&61.02&43.61 \\
        AutoTSP (acc) & \textbf{83.26}&\textbf{79.19}&\textbf{86.89}&\textbf{81.82}&\textbf{57.06}&42.05&\textbf{59.99}&\textbf{42.98}\\
        NHA (acc) & 80.52&69.16&82.17&79.15&56.7&\textbf{47.55}&\textbf{59.99}&37.28\\
        Standard (acc) & 72.49&46.21&77.23&56.88&53.3&29.36&56.65&38.46\\
         \hline
         MOTA (LP) & 97.78 & 95.48 &96.93& 93.35 & 95.62 & 96.6 &81.4& 63.41\\ 
         AutoTSP (LP) & 98.99 & 95.48 &97.26& 93.35 & 99.41 & 98.25 &85.21& 60.19\\ 
         NHA (LP) & 91.96 & 87.14 &94.42& 86.75 & 90.28 & 81.51 &85.21& 64.61\\ 
         Standard (LP) & 80.58 & 52.16 &80.84& 60.77 & 81.05 & 50.84 &80.35& 60.01\\ 
 \hline
 MOTA (LR) & 94.21 & 84.71 &96.32& 87.02 & 88.48 & 68.42 &98.46& 85.11 \\
 AutoTSP (LR) & 88.12 & 84.71 &94.87& 87.02 & 68.85 & 47.8 &92.85& 96.43 \\
 NHA (LR) & 96.35 & 86.06 &92.03& 89.29 & 90.67 & 86.25 &92.85& 64.77 \\
 Standard (LR) & 96.85 & 92.58 &97.92& 97.65 & 97.48 & 95.74 &98.3& 95.37 \\
    \hline
    \end{tabular}
    }
    \label{TSP comparison CNN9}
\end{table}

\begin{figure}[t]
\centering
\begin{subfigure}
  \centering
  \includegraphics[width= 0.255\textwidth]{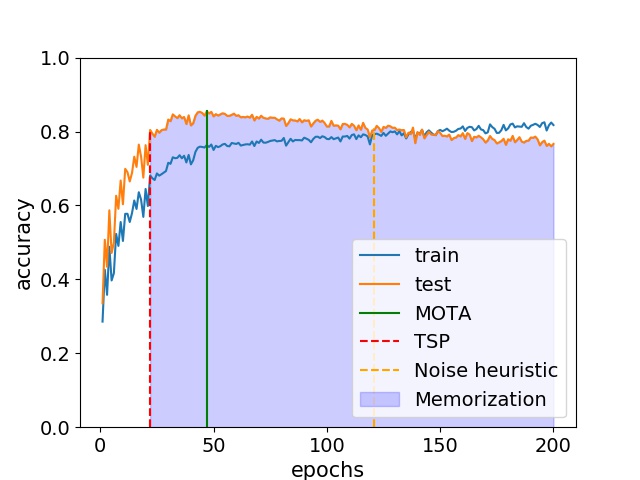}\includegraphics[width= 0.255\textwidth]{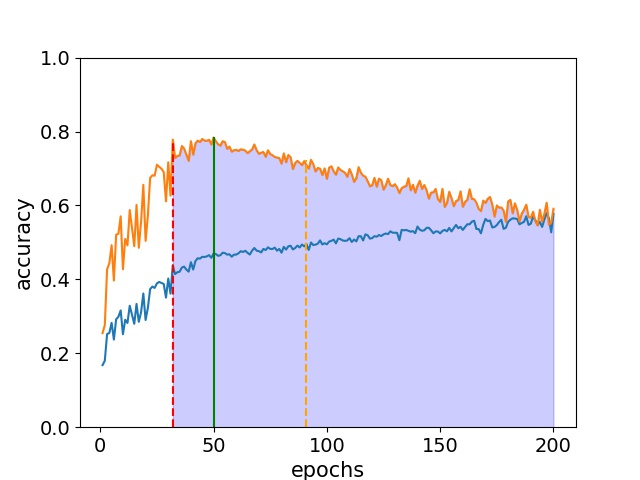}
   
  \caption{AutoTSP comparison on CIFAR-10 with ResNet32: Left 0.2 Sym, Right 0.5 Sym}
  \label{fig: TSP resnet32}
\end{subfigure}
  \vspace{-1\baselineskip}

\end{figure}

We also validated our algorithm with ResNet32 architecture, where the network does not fit all the training samples. Results are presented in Table \ref{TSP comparison ResNet32}. The networks learning drops significantly when fitting on noisy samples and thus the PROCE is close to MOTA in Fig. \ref{fig: Different learning rate architecture NROC PROC CIFAR10 and CIFAR100 sym 0.5}. Similar to CNN9 case, AutoTSP identifies the beginning of memorization region as shown in Fig. \ref{fig: TSP resnet32}. However, we can still observe that the AutoTSP works fairly well on both CIFAR-10 and CIFAR-100, and outperforms NHA on both Sym and Asym cases.

\begin{table}[t]
\caption{Comparison of results with ResNet32 }
    \centering
    \resizebox{\columnwidth}{!}{%
    \begin{tabular}{|c|c|c|c|c|c|c|c|c|}
    \hline
     & \multicolumn{4}{|c|}{\textbf{CIFAR 10}} & \multicolumn{4}{|c|}{\textbf{CIFAR 100}}\\
     & \multicolumn{2}{|c|}{Sym} & \multicolumn{2}{|c|}{Asym}& \multicolumn{2}{|c|}{Sym} & \multicolumn{2}{|c|}{Asym}\\
     &0.2&0.5&0.2&0.4&0.2&0.5&0.2&0.4\\
    \hline
    \hline
        MOTA (acc) &85.45 &78.38&87.22&80.71&56.53&47.35&56.71&42.94 \\
               AutoTSP (acc) & 80.42&\textbf{77.82}&\textbf{86.21}&\textbf{77.09}&\textbf{52.45}&\textbf{45.77}&\textbf{52.73}&33.24\\
        NHA (acc) & \textbf{81}&68.23&81.47&76.47&52.05&36.44&51.87&\textbf{38.61}\\
        Standard (acc) & 76.65&59.09&78.87&56.21&52.05&35.99&51.18&38.1\\
         \hline
         MOTA (LP) & 98.44 & 93.75 &98.41& 91.9 & 99.14 & 96.44 &95.66& 73.27\\
         AutoTSP (LP) & 99.07 & 95.64 &98.87& 90.74 & 99.58 & 97.31 &97.64& 77.67\\ 
         NHA (LP) & 95.01 & 85.93 &93.72& 86.22 & 96.51 & 85.9 &89.63& 69.12\\
         Standard (LP) & 91.61 & 73.33 &88.06& 63.55 & 96.51 & 85.9 &89.56& 66.81\\
         \hline
         MOTA (LR) & 93.83 & 87.83 &94.42& 87.07 & 74.19 & 62.29 &73.09& 61.04 \\ 
         AutoTSP (LR) & 84.3 & 83.58 &91.79& 80.9 & 59.83 & 55.59 &61.37& 37.94 \\ 
         NHA (LR) & 94.71 & 84.73 &92.98& 86.42 & 80.49 & 70.97 &80.34& 67.12 \\ 
         Standard (LR) & 93.55 & 84.36 &94.75& 82.99 & 80.49 & 70.97 &80.48& 69.13 \\ 
    \hline
    \end{tabular}
    }
    \label{TSP comparison ResNet32}
\end{table}

\begin{figure}[t]
\centering
\begin{subfigure}
  \centering
  \includegraphics[width= 0.255\textwidth]{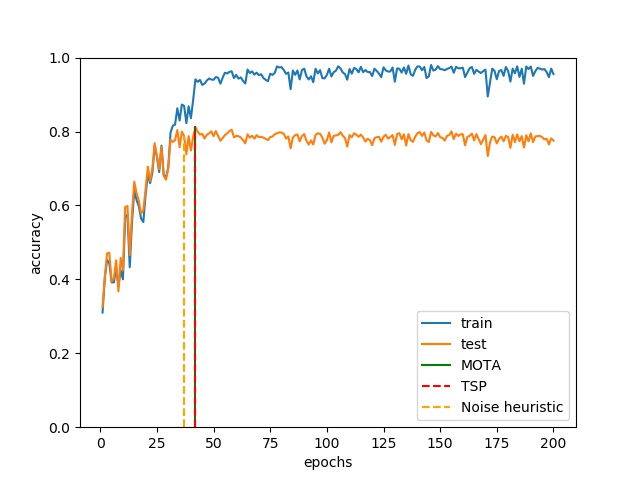}\includegraphics[width= 0.255\textwidth]{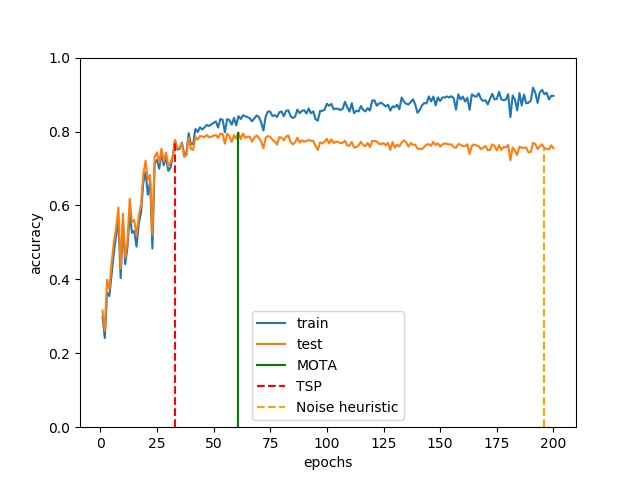}
   
  \caption{AutoTSP comparison on ANIMAL-10N with: Left CNN9, Right ResNet32}
  \label{fig: ANIMAL 10N}
\end{subfigure}
  \vspace{-1\baselineskip}

\end{figure}

\textbf{ANIMAL-10N:} We also conducted experiments on a real world noisy ANIMAL-10N dataset with 50000 training and 5000 testing samples. The noise is caused by human labeling error on 5 pairs of confusing classes and the approximate noise rate is 8\%. It can be seen that AutoTSP still exactly identifies MOTA with CNN9 in the left plot of Fig.\ref{fig: ANIMAL 10N}. With ResNet32 in the right plot, the AutoTSP stops a little earlier than the MOTA, as discussed previously with the results in Table \ref{TSP comparison CNN9}. However, it is still close to MOTA, where the accuracy at MOTA was observed to be 79.6 and AutoTSP accuracy was 78.4. It can also be observed that the NHA stops the training deep in the $SM$ region. The promising results on this real world noisy dataset validates the robustness of AutoTSP.

\begin{figure}[t]
\centering
\begin{subfigure}
  \centering
  \includegraphics[width= 0.255\textwidth]{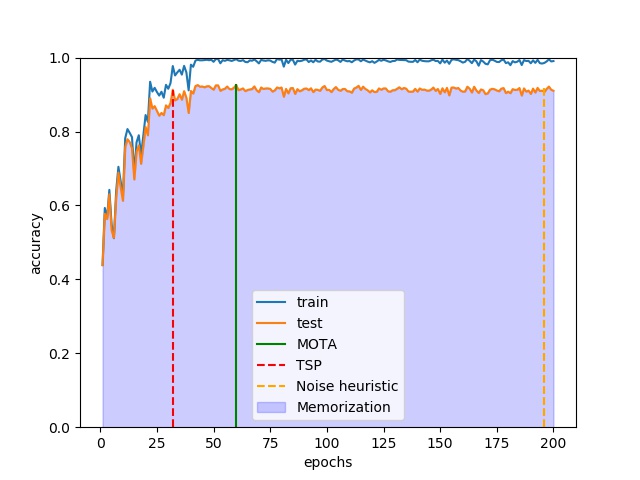}\includegraphics[width= 0.255\textwidth]{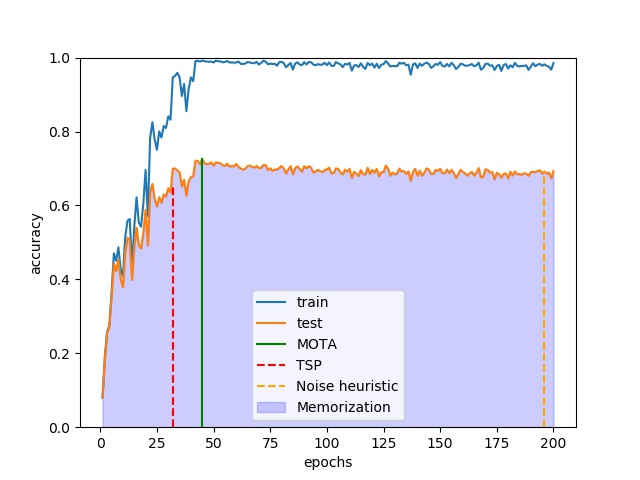}
   
  \caption{AutoTSP comparison with CNN9: Left CIFAR-10, Right CIFAR-100}
  \label{fig: Clean dataset}
\end{subfigure}
  \vspace{-1\baselineskip}

\end{figure}

\textbf{Clean dataset}: We further validated our method on two benchmark clean datasets CIFAR-10 and CIFAR-100 with CNN9 architecture. In general, we utilize a validation dataset and stop the training when the validation accuracy does not improve for a few epochs. Since, utilizing a validation set is not reliable \cite{Sun2019LimitedLabels}, as an alternative, we use the AutoTSP to find the TSP on a clean dataset. It can be observed from Fig. \ref{fig: Clean dataset} that, the AutoTSP is able to  the stop point when the test accuracy no longer improves.
 
\begin{figure}[t]
\centering
\begin{subfigure}
  \centering
  \includegraphics[width= 0.255\textwidth]{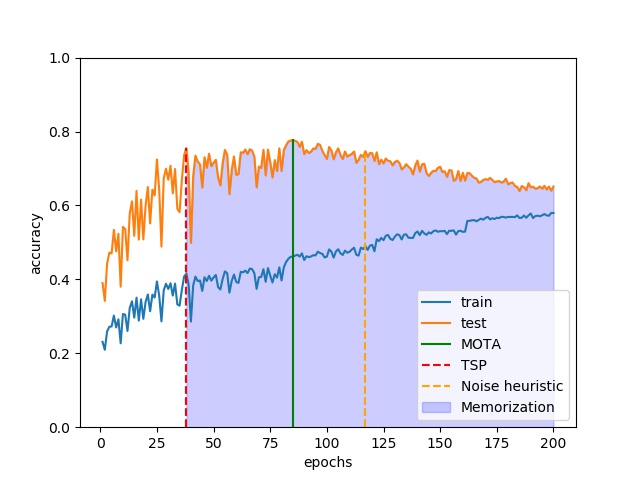}\includegraphics[width= 0.255\textwidth]{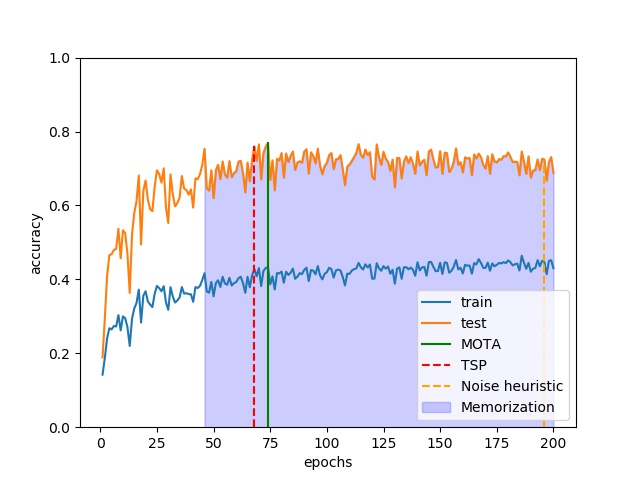}

  \caption{AutoTSP comparison on CIFAR-10 dataset 0.5 Sym on ResNet32:  Left LR2, Right LR3.}
  \label{fig: New Learning rate}
\end{subfigure}
  \vspace{-1\baselineskip}

\end{figure}

\textbf{LR2 and LR3:} Additionally, we trained CIFAR-10 0.5 Sym with ResNet32 using LR2 and LR3 discussed with Fig. \ref{fig: Different learning rate architecture NROC PROC CIFAR10 and CIFAR100 sym 0.5}. It can be observed that the AutoTSP performs well on both cases. AutoTSP finds the beginning of memorization region with the LR2, as can be seen in the left plot if  Fig.\ref{fig: New Learning rate}. The accuracy at MOTA is 77.65 and the accuracy of AutoTSP is 75.53. Similarly, with LR3, it can be observed that the AutoTSP is very close to MOTA in the right plot. These results continue to validate our claim that the rate of change of training accuracy can be utilized to avoid learning on noisy labels.

\subsection{Comparison of test accuracy with other baseline methods:}

For additional validation, we concatenate AutoTSP with the INCV and co-teaching method (referred as AutoTSP+INCV in Table \ref{tab:concatnate}) to obtain higher test accuracy. We compare the test accuracy with the following baselines: F-correction, Decoupling, Coteaching, MentorNet, D2L, INCV and the standard training method. Additionally, we also compare with the test accuracy obtained when trained only on the clean set of the noisy training data (Labeled as Train on clean in the table). The results are shown in Table \ref{tab:concatnate}.

The objective of AutoTSP is to find the best training stop point close to the MOTA. Thus, it is not fair to compare AutoTSP directly with the other baseline methods, whose objective is to improve the test accuracy by re-labeling the noisy samples or, selecting higher number of samples iteratively, etc. We utilize the INCV discard measure to remove a percentage of high-loss values at $e_{tsp}$ which are believed to be noisy samples. Then, we retrain on the remaining samples, find the $e_{tsp}$ and discard few more high-loss samples. So, the new discarded dataset will be less noisy than the given training dataset. Now, we train the network using co-teaching method instead of standard-training, which proved to be effective with noisy labels.

Since, INCV is the highest among all the other baseline methods in Table \ref{tab:concatnate}, we compare our AutoTSP+INCV results only with the INCV. For CIFAR-10, we can see that our AutoTSP + INCV and INCV accuracy are very close. The significance of AutoTSP can be observed for CIFAR-100 as our accuracy is higher than INCV. This is because, INCV depends largely on the noise estimation and the noise estimation becomes inaccurate for higher number of classes. On the other hand, we find the TSP without using the noise information. We only utilized the noise information to adopt the INCV discard measure.

The claims in Table \ref{tab:concatnate} are supported by Table \ref{tab:lp and lr second round}. For CIFAR-10 Sym, AutoTSP+INCV and INCV LP and LR are very similar. The significance of AutoTSP can be observed with Asym 0.4 on CIFAR-10, as the LP and LR of AutoTSP+INCV is higher. On CIFAR-100 it can be seen that the LR of AutoTSP+INCV is much less compared to INCV. This is because, the AutoTSP would stop the training before several noisy labels are memorized as discussed earlier. Thus, high LP is achieved for both symmetric noise cases on CIFAR-100.

\begin{table}[t]
\caption{Test accuracy comparison with other baseline methods with ResNet32, *the values are taken from the INCV paper and the highest test accuracy is in bold.}
    \centering
    \resizebox{\columnwidth}{!}{%

    \begin{tabular}{|p{4cm}|c|c|c|c|c|}
    \hline
     & \multicolumn{3}{|c|}{\textbf{CIFAR 10}} & \multicolumn{2}{|c|}{\textbf{CIFAR 100}}\\
     & \multicolumn{2}{|c|}{Symmetric} & \multicolumn{1}{|c|}{Asymmetric}& \multicolumn{2}{|c|}{Symmetric} \\
     &0.2&0.5&0.4&0.2&0.5\\
    \hline
    \hline
        Standard training & 85 & 75 & 66 &57 &43\\ 
 \hline
 F-correction \cite{Patrini2017MakingApproach}& 85.08* & 79.3* & 83.55* &-&- \\ 
 \hline
 Decoupling \cite{Malach2017DecouplingUpdate}& 86.72*&79.31* &75.27* &-&-\\
 \hline
 Co-teaching \cite{Han2018Co-teaching:Labels}&89.15 &82.23 & 83.95&57&51\\
 \hline
 MentorNet \cite{Jiang2018MentorNet:Labels}&88.36* &77.10* & 77.33*&-&-\\
 \hline
 D2L \cite{Ma2018Dimensionality-DrivenLabels}&86.12*&67.39* &85.57* &-&-\\
 \hline
 INCV \cite{Chen2019UnderstandingLabels}&\textbf{89.65} & \textbf{84.81} & 85.84&60.4&53\\
 \hline
 \textbf{AutoTSP+INCV} & 89.49 & 84.5& \textbf{86.34} &\textbf{62.5}&\textbf{53.8}\\
 \hline
 Train on clean & 90& 89.4& 89&64&58\\
 \hline
    \end{tabular}
    }
    \label{tab:concatnate}
\end{table}

\begin{table}[ht]
\caption{Comparison of LP and LR of AutoTSP+INCV and INCV with ResNet32}
    \centering
    \resizebox{\columnwidth}{!}{%
    \begin{tabular}{|p{4cm}|c|c|c|c|c|c|c|c|}
    \hline
     & \multicolumn{4}{|c|}{\textbf{CIFAR 10}} & \multicolumn{4}{|c|}{\textbf{CIFAR 100}}\\
     & \multicolumn{2}{|c|}{Symmetric} & \multicolumn{2}{|c|}{Asymmetric}& \multicolumn{2}{|c|}{Symmetric} & \multicolumn{2}{|c|}{Asymmetric}\\
     &0.2&0.5&0.2&0.4&0.2&0.5&0.2&0.4\\
    \hline
    \hline
    LP (INCV) & 0.98 & 0.92 &0.97& 0.82 & 0.98 & 0.97&0.96& 0.73\\ 
 
 LR (INCV) & 0.93 & 0.89 &0.95& 0.92 & 0.93& 0.63&0.73&0.65 \\ 
 \hline
        LP (AutoTSP+INCV) & 0.98 & 0.95 &0.98& 0.89 & 0.99 & 0.96 &0.95& 0.71\\ 
 
 LR (AutoTSP+INCV) & 0.95 & 0.85 &0.93& 0.94 & 0.73 & 0.64 &0.74& 0.57 \\ 

 \hline
  
  \end{tabular}
  }
    
    \label{tab:lp and lr second round}
\end{table}

\section{Conclusion}
 
In this work, we have shown that the rate of change of training accuracy can be essential in understanding the behavior of learning on noisy labels. Our key idea is to monitor the smoothness in the training accuracy to identify the training stop region. We further proposed a heuristic algorithm to automatically identify a training stop point based on the small-learning assumption. Our algorithm does not require a clean validation set and does not depend on noise estimation or the type of noise. We conducted several experiments to validate the robustness of our algorithm. One drawback of early stopping, in general, is that the network does not train on the harder-clean samples learned in the later stages, which could improve the networks generalization performance. Therefore, one possible future work is to discard noisy samples effectively without depending on the noise estimation.

\bibliographystyle{IEEEtranS}
\bibliography{references}

\newpage

\section{Supplementary material}

   \subsection{Choice of $\beta_{1}$:}
   To find the training stop region, we monitor the magnitude of PROCE and NROCE over different intervals and identify the point at which the magnitude decreases. Since, we are looking to identify the reduction of the rate of change at different intervals in the training process, a few smaller magnitude epochs in the same interval could result in finding a region too early. Thus, to ignore such an effect, we choose a large interval size to identify the consistent drop in magnitude. We also consider three different interval sizes for redundancy and compare the reduction point among the three intervals to choose the training stop region. We choose $\beta_{1}$ to be 3, 4 and 5 for our experiments and denote this case as \{3, 4, 5\}. Now, we consider the region between the \textbf{min}(min(NROCE\{3, 4, 5\}), max(PROCE\{3, 4, 5\})) and \textbf{max}(max(NROCE\{3, 4, 5\}), max(PROCE\{3, 4, 5\})) as the training stop region.
   
   Table \ref{TSP comparison ResNet110 different interval} shows the accuracy results when we set $\beta_{1}$ to \{5, 6, 7\}. It can be observed that the accuracy is similar in both cases. This shows that the algorithm is not very sensitive to different interval sizes. This result makes qualitative sense, because, the reduction happens at the same point in the training irrespective of the different interval sizes.
   
   \begin{table}[ht]
\caption{Comparison of test accuracy at the TSP with different interval sizes on ResNet110 }
    \centering
    \resizebox{\columnwidth}{!}{%

    \begin{tabular}{|c|c|c|c|c|c|c|c|c|}
    \hline
     & \multicolumn{4}{|c|}{\textbf{CIFAR 10}} & \multicolumn{4}{|c|}{\textbf{CIFAR 100}}\\
     & \multicolumn{2}{|c|}{Symmetric} & \multicolumn{2}{|c|}{Asymmetric}& \multicolumn{2}{|c|}{Symmetric} & \multicolumn{2}{|c|}{Asymmetric}\\
     &0.2&0.5&0.2&0.4&0.2&0.5&0.2&0.4\\
    \hline
    \hline
        
          AutoTSP ($\beta_{1}$ \{3, 4, 5\} acc) & \textbf{85.01}&\textbf{75.86}&\textbf{88.28}&\textbf{76.31}&\textbf{57.33}&\textbf{46.65}&56.03&\textbf{42.25}\\
        AutoTSP ($\beta_{1}$ \{5, 6, 7\} acc) & 83.67&\textbf{75.86}&\textbf{88.28}&\textbf{76.31}&56.35&\textbf{46.65}&\textbf{57.99}&\textbf{42.25}\\

    \hline
    \end{tabular}
    }
    \label{TSP comparison ResNet110 different interval}
\end{table}

\subsection{Small-learning assumption:}

The small-learning assumption negates the magnitude when $MT^(e)$ is less than the threshold $\theta_{1}$. Since, in the initial stages, there would be only a few SLEs, the following epochs would be able to compensate for the drop and achieve positive sum, meaning achieving higher test accuracy. But, in the later stages, there would be higher number of SLEs and fewer epochs with high magnitude. So, these fewer high magnitude epochs cannot overcome the large negative cumulative sum. Therefore, this results in several zeros in the later stages. Thus, we can claim that our small-learning assumption is strong. However, we cannot use this assumption alone to find the TSP. As the rate tends to increase slightly in the SM region, there could be a few false positives which indicate the test accuracy is increasing at the end. Thus, to avoid these false positives, we only apply the small-learning assumption within the training stop region, assuming the training stop region truncates this SM region as shown in Fig \ref{fig: LR clean and noise}, \ref{fig: NROC PROC CIFAR10 sym asym 0.2} \ref{fig: Different architecture NROC PROC CIFAR10 and CIFAR100 sym 0.5}.

 \subsection{Choice of $\theta_{1}$ and $\theta_{2}$:}

To show the impact of the thresholds, we set $\theta_{1}$ to 0.2, 0.5, 0.8 and $\theta_{2}$ to 0, 0.5. Note that, if $\theta_{1}$ is set to zero the small-learning assumption is ignored. 

Table \ref{TSP comparison different theta ResNet110} shows accuracy results for different $\theta_{1}$ values. It can be observed that the results are similar for the three cases, showing that the algorithm is not very sensitive to the thresholds chosen. However, if the threshold is set even higher, it found a stop point too early in the training. So, the threshold size should increase with the epoch number, as, in the later stages, even the higher magnitude epochs result in learning on noisy labels. But, finding that relation is harder without knowing the noise rate. Both INCV \cite{Chen2019UnderstandingLabels} and co-teaching \cite{Han2018Co-teaching:Labels} utilize a threshold to select a percentage of loss values as small or high loss values. However, the two methods utilize the noise ratio in their thresholds. But our algorithm assumes that the noise rate is unknown. Thus, we observed the rate of change across different datasets, architectures and hyper-parameters to choose the threshold. Based on the observation, we set the ideal $\theta_{1}$ to be $0.5$.

The results are similar for both $\theta_{2}$ values 0 and 0.5. $\theta_{2}$ helps when the training stop region finds an upper bound in the later region by further penalizing the small-magnitude epochs. This, $\theta_{2}$ can be ignored if threshold value is not kept constant and increased with the epoch number.

\begin{table}[ht]
\caption{Comparison results of test accuracy, LP and LR with ResNet110 for different $\theta_{1}$ values when $\theta_{2}$ is set to 0.5}
    \centering
    \resizebox{\columnwidth}{!}{%

    \begin{tabular}{|c|c|c|c|c|c|c|c|c|}
    \hline
     & \multicolumn{4}{|c|}{\textbf{CIFAR 10}} & \multicolumn{4}{|c|}{\textbf{CIFAR 100}}\\
     & \multicolumn{2}{|c|}{Symmetric} & \multicolumn{2}{|c|}{Asymmetric}& \multicolumn{2}{|c|}{Symmetric} & \multicolumn{2}{|c|}{Asymmetric}\\
     &0.2&0.5&0.2&0.4&0.2&0.5&0.2&0.4\\
    \hline
    \hline
     AutoTSP ($\theta_{1}$=0.2 acc) & 83.86&74.76&84.85&\textbf{76.36}&57.33&\textbf{46.65}&56.03&\textbf{42.25}\\
      
        AutoTSP ($\theta_{1}$=0.5 acc) & \textbf{85.01}&\textbf{75.86}&\textbf{88.28}&76.31&57.33&\textbf{46.65}&\textbf{56.43}&\textbf{42.25}\\
        
           AutoTSP ($\theta_{1}$=0.8 acc) & 83.7&\textbf{75.86}&84.85&76.31&\textbf{58.04}&\textbf{46.65}&56.03&\textbf{42.25}\\
      
        \hline
        AutoTSP ($\theta_{1}$=0.2  LP) & 98.11 & 93.57 &98.42& 88.1 & 94.5 & 96.45 &87.24& 63.1\\ 
        
        AutoTSP ($\theta_{1}$=0.5 LP) & 97.6 & 95.2 &97.54& 87.54 & 94.5 & 96.45 &84.25& 63.1\\ 
       
       AutoTSP ($\theta_{1}$=0.8 LP) & 98.27 & 95.2 &98.42& 87.54 & 95.61 & 97.16 &84.25& 63.1\\  
      
 \hline
 AutoTSP ($\theta_{1}$=0.2 LR)  & 92.47 & 82.72 &91.12& 83.56 & 88.9 & 61.48 &89.46& 88.21 \\
 
  AutoTSP ($\theta_{1}$=0.5 LR)  & 94.23 & 80.23 &96.56& 82.66 & 88.9 & 61.48 &93.83& 88.21 \\

    AutoTSP ($\theta_{1}$=0.8 LR)  & 91.42 & 80.23 &91.12& 82.66 & 88.08 & 60.9 &93.83& 88.21 \\
 
    \hline
    \end{tabular}
    }
    \label{TSP comparison different theta ResNet110}
\end{table}

\subsection{Choosing a different training stop point:}

Table \ref{TSP comparison ResNet110 nonzero last}, \ref{TSP comparison CNN9 last nonzero} and \ref{TSP comparison ResNet32 last nonzero} show the results for different architectures when a different training stop point is chosen within the training stop region. We implemented the last non-zero element within the training stop region as the training stop point instead of the maximum cumulative sum (argmax). However,  we observed that argmax performs slightly better than the last non-zero element.

\begin{table}[ht]
\caption{Comparison results of test accuracy, LP and LR with ResNet110 for different training stop points}
    \centering
    \resizebox{\columnwidth}{!}{%

    \begin{tabular}{|c|c|c|c|c|c|c|c|c|}
    \hline
     & \multicolumn{4}{|c|}{\textbf{CIFAR 10}} & \multicolumn{4}{|c|}{\textbf{CIFAR 100}}\\
     & \multicolumn{2}{|c|}{Symmetric} & \multicolumn{2}{|c|}{Asymmetric}& \multicolumn{2}{|c|}{Symmetric} & \multicolumn{2}{|c|}{Asymmetric}\\
     &0.2&0.5&0.2&0.4&0.2&0.5&0.2&0.4\\
    \hline
    \hline
     
        AutoTSP (max acc) & \textbf{85.01}&\textbf{75.86}&\textbf{88.28}&\textbf{76.31}&\textbf{57.33}&\textbf{46.65}&\textbf{56.03}&\textbf{42.25}\\
  
        AutoTSP (last non-zero acc) & 84.7&74.47&86.07&\textbf{76.31}&\textbf{57.33}&\textbf{46.65}&\textbf{56.03}&\textbf{42.25}\\

        \hline
          AutoTSP (max LP) & 97.6 & 95.2 &97.54& 87.54 & 94.5 & 96.45 &84.25& 63.1\\ 
        AutoTSP (last non-zero LP) & 98.27 & 93.94 &95.64& 87.54 & 94.5 & 97.16 &84.25& 63.1\\ 
      
 \hline
 AutoTSP (max LR)  & 94.23 & 80.23 &96.56& 82.66 & 88.9 & 61.48 &93.83& 88.21 \\
  AutoTSP (last non-zero LR)  & 91.42 & 82.39 &97.04& 82.66 & 88.9 & 60. &93.83& 88.21 \\

    \hline
    \end{tabular}
    }
    \label{TSP comparison ResNet110 nonzero last}
\end{table}

\begin{table}[ht]
\caption{Comparsion results of test accuracy, LP and LR with CNN9 for different training stop points}
    \centering
    \resizebox{\columnwidth}{!}{%

    \begin{tabular}{|c|c|c|c|c|c|c|c|c|}
    \hline
     & \multicolumn{4}{|c|}{\textbf{CIFAR 10}} & \multicolumn{4}{|c|}{\textbf{CIFAR 100}}\\
     & \multicolumn{2}{|c|}{Symmetric} & \multicolumn{2}{|c|}{Asymmetric}& \multicolumn{2}{|c|}{Symmetric} & \multicolumn{2}{|c|}{Asymmetric}\\
     &0.2&0.5&0.2&0.4&0.2&0.5&0.2&0.4\\
    \hline
    \hline
        AutoTSP (max acc) & \textbf{83.26}&\textbf{79.19}&86.89&\textbf{81.82}&\textbf{57.06}&42.05&\textbf{59.99}&\textbf{42.98}\\
     
        AutoTSP (last non-zero acc) & \textbf{83.26}&77.64&\textbf{87.58}&79.15&56.91&\textbf{42.43}&58.36&42.61\\
    
         \hline
        AutoTSP (max LP) & 98.99 & 95.48 &97.26& 93.35 & 99.41 & 98.25 &85.21& 60.19\\ 
         AutoTSP (last non-zero LP) & 98.99 & 93.95 &96.93& 86.75 & 98.62 & 97.62 &84& 60.35\\ 
         
 \hline
AutoTSP (max LR) & 88.12 & 84.71 &94.87& 87.02 & 68.85 & 47.8 &92.85& 96.43 \\
 AutoTSP (last non-zero LR) & 88.12 & 87.17 &96.32& 89.29 & 75.5 & 50.29 &93.1& 96.89 \\
 
    \hline
    \end{tabular}
    }
    \label{TSP comparison CNN9 last nonzero}
\end{table}

\begin{table}[ht]
\caption{Comparsion results of test accuracy, LP and LR with ResNet32 for different training stop points}
    \centering
    \resizebox{\columnwidth}{!}{%
    \begin{tabular}{|c|c|c|c|c|c|c|c|c|}
    \hline
     & \multicolumn{4}{|c|}{\textbf{CIFAR 10}} & \multicolumn{4}{|c|}{\textbf{CIFAR 100}}\\
     & \multicolumn{2}{|c|}{Sym} & \multicolumn{2}{|c|}{Asym}& \multicolumn{2}{|c|}{Sym} & \multicolumn{2}{|c|}{Asym}\\
     &0.2&0.5&0.2&0.4&0.2&0.5&0.2&0.4\\
    \hline
    \hline
          AutoTSP (max acc) & \textbf{80.42}&\textbf{77.82}&\textbf{86.21}&\textbf{77.09}&52.45&\textbf{45.77}&52.73&33.24\\
              AutoTSP (acc) & \textbf{80.42}&77.39&\textbf{86.21}&\textbf{77.09}&\textbf{54.64}&\textbf{45.77}&\textbf{54}&\textbf{36.85}\\
       
         \hline
        AutoTSP (max LP) & 99.07 & 95.64 &98.87& 90.74 & 99.58 & 97.31 &97.64& 77.67\\
         AutoTSP (LP) & 99.07 & 94.97 &98.87& 90.74 & 99.51 & 97.31 &96.8& 77.08\\ 
       
         \hline
         AutoTSP (max LR) & 84.3 & 83.58 &91.79& 80.9 & 59.83 & 55.59 &61.37& 37.94 \\ 
         AutoTSP (LR) & 84.3 & 84.56 &91.79& 80.9 & 67.09 & 55.59 &66.64& 44.1 \\ 
       
    \hline
    \end{tabular}
    }
    \label{TSP comparison ResNet32 last nonzero}
\end{table}

\subsection{Architecture description:}

The 9-layer CNN architecture we used is described below in Table \ref{CNN9}.

\begin{table}[ht]
\caption{9 layer CNN architecture}
    \begin{center}
    \begin{tabular}{||c||} 
    \hline
    \textbf{CNN-9}  \\ [0.5ex] 
    \hline\hline
    3x3 conv,128 ReLu \\ 
    Batchnorm\\
    3x3 conv,128 ReLu \\ 
    Batchnorm\\
    3x3 conv,128 ReLu \\ 
    Batchnorm\\
    2x2 maxpool\\
    \hline
    3x3 conv,256 ReLu \\ 
    Batchnorm\\
    3x3 conv,256 ReLu \\ 
    Batchnorm\\
    3x3 conv,256 ReLu \\ 
    Batchnorm\\
    2x2 maxpool\\
    \hline
    3x3 conv,512 ReLu \\ 
    Batchnorm\\
    3x3 conv,512 ReLu \\ 
    Batchnorm\\
    3x3 conv,512 ReLu \\ 
    Batchnorm\\
    \hline
    average pool\\
    \hline
    Fully connected layer\\
    
    \hline
    
    \end{tabular}
    \end{center}
    \label{CNN9}
\end{table}

\end{document}